\documentclass[preprint,12pt,authoryear]{elsarticle}

\usepackage{amssymb}
\usepackage{amsmath}
\usepackage{commath}
\usepackage{caption}
\usepackage{array}
\usepackage{booktabs}
\usepackage{multirow}
\usepackage{subcaption}
\usepackage{arydshln}
\usepackage{latexsym}
\usepackage{makecell}
\usepackage{tikz}
\usepackage{xcolor}
\usepackage{pgfplots} 
\pgfplotsset{compat=1.18} 

\definecolor{input}{HTML}{9FC5E8}
\definecolor{encoder}{HTML}{FFF2CC}
\definecolor{decoder}{HTML}{F4CCCC}
\definecolor{bottleneck}{HTML}{C6F3A4}
\definecolor{output}{HTML}{F9CB9C}

\tikzset{input/.style={black,draw=black,fill=input,rectangle,minimum height=0.8cm}}
\tikzset{encoder/.style={black,draw=black,fill=encoder,rectangle,minimum height=0.8cm}}
\tikzset{bottleneck/.style={black,draw=black,fill=bottleneck,rectangle,minimum height=0.8cm}}
\tikzset{decoder/.style={black,draw=black,fill=decoder,rectangle,minimum height=0.8cm}}
\tikzset{output/.style={black,draw=black,fill=output,rectangle,minimum height=0.8cm}}

\begin{document}

\begin{frontmatter}

\title{Compression Method for Solar Polarization Spectra Collected from Hinode SOT/SP Observations}

\author[1]{Jargalmaa Batmunkh}
\ead{f22l006h@mail.cc.niigata-u.ac.jp}

\author[1]{Yusuke Iida}
\ead{iida@ie.niigata-u.ac.jp}

\author[2]{Takayoshi Oba}
\ead{oba@mps.mpg.de}

\author[3]{Haruhisa Iijima}
\ead{h.iijima@isee.nagoya-u.ac.jp}

\affiliation[1]{organization={Niigata University},
            addressline={8050 Ikarashi 2-no-cho, Nishi-ku}, 
            city={Niigata},
            postcode={950-2181}, 
            state={Niigata},
            country={Japan}}

\affiliation[2]{organization={Max Planck Institute for Solar System Research},
            addressline={Justus-von-Liebig-Weg 3}, 
            city={Göttingen},
            postcode={37077}, 
            country={Germany}}

\affiliation[3]{organization={Nagoya University},
            addressline={Furo-cho, Chikusa-ku}, 
            city={Nagoya},
            postcode={464-8601}, 
            state={Aichi},
            country={Japan}}

\begin{abstract}

The complex structure and extensive details of solar spectral data, combined with a recent surge in volume, present significant processing challenges. To address this, we propose a deep learning-based compression technique using deep autoencoder (DAE) and 1D-convolutional autoencoder (CAE) models developed with Hinode SOT/SP data. We focused on compressing Stokes I and V polarization spectra from the quiet Sun, as well as from active regions, providing a novel insight into comprehensive spectral analysis by incorporating spectra from extreme magnetic fields. The results indicate that the CAE model outperforms the DAE model in reconstructing Stokes profiles, demonstrating greater robustness and achieving reconstruction errors around the observational noise level. The proposed method has proven effective in compressing Stokes I and V spectra from both the quiet Sun and active regions, highlighting its potential for impactful applications in solar spectral analysis, such as detection of unusual spectral signals.

\end{abstract}


\end{frontmatter}

\section{Introduction}
\label{sec_introduction}
Observational spectral data encapsulates important and varied physical information with a multi-dimensional structure about astronomical bodies, necessitating thorough investigation and analysis for a comprehensive understanding of space. The increase in the number of observatory instruments in recent years has led to a substantial growth in the volume of astronomical data. This surge not only emphasizes the significance of studying such data but also opens up promising opportunities for leveraging deep learning techniques in the processing and analysis of these vast datasets in the big data era. One approach to handling such intricate data is the feature extraction technique, which takes the high-dimensional raw data as input, compresses it, and reconstructs it to the original size.
The most important features of the original high-dimensional data are extracted in the compressed part, enabling them to serve as representatives of the original complex dataset in subsequent studies. Autoencoders \citep{lecun1987phd, Kramer1991-zs, Goodfellow-et-al-2016} play a powerful role in deep learning-based dimensionality reduction. Through this compression approach, further studies such as anomaly detection \citep{8363930, RYU2023475} and classification \citep{8524276, 10.1145/3426020.3426093} of data can also be accomplished. 

In the latter part of the 2010s, several studies aimed to develop and apply compression methods for spectral data, particularly in the context of galaxy observations. \citet{Portillo2020} utilized variational autoencoder models to compress galaxy spectra by reducing it to six parameters, offering more accurate reconstructions than principal component analysis (PCA). \citet{Melchior_2023} introduced an architecture to represent and generate restframe galaxy spectra from 6 to 10 latent parameters, resulting in accurate reconstructions with superresolution and reduced noise.

When compared to data in other fields of space science, solar spectral data stand out in terms of their increased precision and complexity in higher dimensionality, encompassing details about light polarization, temperature, and the magnetic field on the solar surface. Therefore, processing this type of observational data poses a significant challenge. Previous works referring to the representational dimension of solar polarimetric spectra include, \citet{2006ApJ...646.1445A}'s two-part minimum description length principle for approximation model selection, which suggests the optimal eigenvector dimension for denoising PCA, and \citet{Asensio_Ramos_2007}'s intrinsic dimensionality estimation method for spectropolarimetry data. \citet{2002ApJ...575..529L} implemented a PCA inversion technique using 10 eigenprofiles for a single Stokes profile. A feature extraction technique by \citet{SocasNavarro2005} for simulated solar profiles, based on a multi-layer perceptron, represents one of the first uses of a neural network for solar spectra, achieving higher accuracy than previous methods such as PCA but requiring significant computational expense. Studies conducted on inversion techniques using deep learning, include \citet{refId0}'s convolutional neural network-based inversion method for Stokes profiles using Hinode \citep{Kosugi2007-db} data. Additionally, \citet{AsensioRamos2019} introduced convolutional neural networks that output thermodynamic and magnetic properties from synthetic Stokes profiles, and achieved a precision comparable to the standard technique. Regarding deep learning-based solar spectral compression, \citet{9461879} used a fully connected autoencoder to reduce one-dimensional quiet Sun spectra, collected by NASA's IRIS \citep{2014SoPh..289.2733D} satellite, from 110 to 4 in size, achieving an average reconstruction error comparable to the variations in the line continuum.

Upon reviewing previous works, it becomes apparent that compression techniques for observational solar spectra have primarily been developed for one-dimensional spectra related to spatial positions in the quiet Sun. However, active regions cannot be disregarded, as they are associated with a variety of significant solar phenomena—such as solar flares, solar jets, and coronal mass ejections—necessitating thorough study as important regions of interest. This motivates our proposal to develop an efficient compression method for solar polarization spectra, applicable to both the quiet Sun and active regions, by utilizing two-dimensional key polarimetric parameters.

We conduct our study using observational solar spectra from Hinode SOT/SP \citep{Tsuneta2008, Suematsu2008, Lites2013}, a collaborative mission of JAXA, NASA, and ESA. This mission has been collecting solar spectro-polarimetric data since 2006, constituting an extensive solar spectral database suitable for our work. Our study introduces the compression of solar spectra through the development of two distinct models: a deep autoencoder (DAE) and a 1D-convolutional autoencoder (CAE). Considering the intricate nature of Stokes profiles characterized by high noise levels, we exclusively focus on Stokes I (total intensity) and Stokes V (circular polarization) selected from the set of four parameters.

In Section \ref{sec_hinode_data}, we provide a description of the Hinode data, followed in Section \ref{sec_compression_model} by a comprehensive explanation of the methods applied in our study. Sections \ref{sec_results} and \ref{sec_discussion} present the results and discussion, respectively. In Section \ref{sec_conclusion}, we conclude the paper with a brief summary.

\section{Hinode SOT/SP Data}
\label{sec_hinode_data}
The Solar Optical Telescope (SOT) on Hinode is equipped with a spectro-polarimeter (SP) that observes the sun, capturing high-resolution spatial images and spectro-polarimetric data. Its long-term collection of solar spectral data over more than 15 years has enabled us to utilize it in the development of our deep learning-based approach.

\begin{figure}[htb]
    \centering
    \captionsetup{font=scriptsize, justification=centering}
      \includegraphics[clip,width=0.95\textwidth]{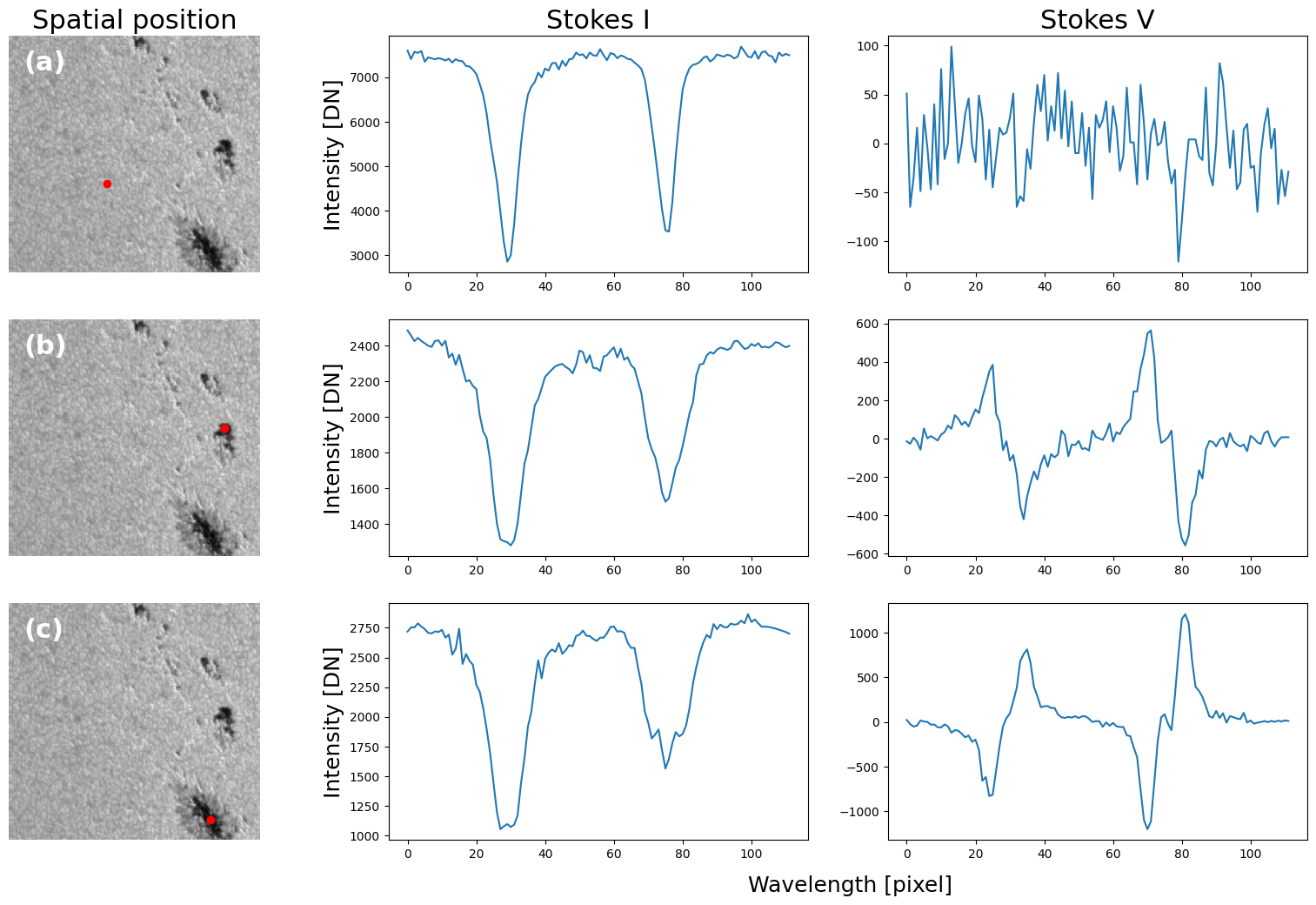}
    \caption{Sample profiles of Stokes parameters corresponding to spatial positions marked in red are provided for (a) quiet Sun, (b) pore, and (c) sunspot core in the FoV image. 
    }
    \label{fig:spatial_pos_abc_IV}
\end{figure}

Hinode SOT/SP data consists of spatio-temporal spectro-polarimetric information covering Fe I line pair profiles at 630.15 and 630.25 nm, along with their nearby continuums. A sampling slit with a width of ~0.15" was used to construct these line pair profiles. The data dimension is 2D-space×1D-wavelength×1D-polarimetry. The spatial, wavelength, and polarimetry dimensions correspond to different fields of views (FoV), 112 wavelength points, and four Stokes parameters (I, Q, U, and V). The profiles of the Stokes I and V exhibit distinguishable noise levels at various spatial positions, as shown in Fig. \ref{fig:spatial_pos_abc_IV}. Notably, in various spatial positions with varying magnetic fields, Stokes I exhibits a smoother profile than Stokes V.

We selected Hinode/SP Level 1 data \citep{Lites2013-fa} observed on 2011-09-25 at the timestamp 20:01:04, downloaded from the Community Spectropolarimetric Analysis Center (CSAC, \citeyear{Community_Spectropolarimetric_Analysis_Center_CSAC2006-zx}) website, based on its capture near the center of the solar disk containing both sunspots and quiet Sun regions. 
The data consists of FITS files of individual scans, each with a FoV in the y and x directions of 162.304" and 0.295", respectively. After combining the FITS file scans along the x direction, we reconstructed a 2D spectro-polarimetric (SP) image with a size of 162.304" in the y direction and 151.142" in the x direction. To isolate individual spectral data (each pixel of the FoV) while disregarding spatial information, we transformed the 2D-space into a 1D-pixel dimension, resulting in a new structure of 1D-pixel×1D-wavelength×1D-polarimetry. 

\section{Compression Model}
\label{sec_compression_model}

\subsection{Autoencoder}
\label{subsec_autoencoder}

The autoencoder is recognized as one of the most notable representatives of neural network-based feature extraction approaches. Its architecture contains encoding and decoding components, each comprised of neural network layers working together to efficiently reduce the size of the input through reconstruction. This encoder-decoder structured dimensionality reduction technique works effectively on non-linearly connected data. Furthermore, it contributes to reducing noise within the data \citep{Saura2023-vm}, potentially leading to reconstructed spectra with decreased observational and instrumental noise. This characteristic positions it as a strong candidate for model selection. The primary goal of the autoencoder is to reconstruct input data into an output that closely resembles the input. The encoder decreases the dimension of the input, while the decoder performs the reverse operation, increasing the lower dimensional input back to the size of the original input. This lower-dimensional bottleneck, known as the feature vector, serves as the compressed representation of the original input. 

\begin{figure}[htb!]
  \centering
  \begin{subfigure}[c]{.465\textwidth}
    \centering
    \resizebox{\textwidth}{!}{%
      \begin{tikzpicture}
    \node[input,rotate=90,minimum width=4.2cm] (input) at (1.25,0) {\large$\textbf{Input spectra}$};
    \node[encoder,rotate=90,minimum width=5cm] (fc1) at (2.5,0) {\large$\text{Dense}_{448}$\,+\,$\text{BN}$\,+\,$\text{ELU}$};
    \node[encoder,rotate=90,minimum width=5cm] (fc2) at (3.75,0) {\large$\text{Dense}_{224}$\,+\,$\text{BN}$\,+\,$\text{ELU}$};
    \node[encoder,rotate=90,minimum width=5cm] (fc3) at (5,0) {\large$\text{Dense}_{112}$\,+\,$\text{BN}$\,+\,$\text{ELU}$};
    
    \node[bottleneck,rotate=90,minimum width=2.5cm] (fc4) at (6.25,0) {\large$\text{Dense}$};
    
    \node[decoder,rotate=90,minimum width=5cm] (fc5) at (7.5,0) {\large$\text{Dense}_{112}$\,+\,$\text{BN}$\,+\,$\text{ELU}$};
    \node[decoder,rotate=90,minimum width=5cm] (fc6) at (8.75,0) {\large$\text{Dense}_{224}$\,+\,$\text{BN}$\,+\,$\text{ELU}$};
    \node[decoder,rotate=90,minimum width=5cm] (fc7) at (10,0) 
    {\large$\text{Dense}_{448}$\,+\,$\text{BN}$\,+\,$\text{ELU}$};
    \node[output,rotate=90,minimum width=4.2cm] (output) at (11.25,0) {\large$\textbf{Reconstruction}$};
    
    \draw[->] (input) -- (fc1);
    \draw[->] (fc1) -- (fc2);
    \draw[->] (fc2) -- (fc3);
    \draw[->] (fc3) -- (fc4);
    \draw[->] (fc4) -- (fc5);
    \draw[->] (fc5) -- (fc6);
    \draw[->] (fc6) -- (fc7);
    \draw[->] (fc7) -- (output);

  \end{tikzpicture}
  }
  \caption{DAE}
  \end{subfigure}\quad\quad
  \begin{subfigure}[c]{.465\textwidth}
  \resizebox{\textwidth}{!}{%
    \begin{tikzpicture}
    \node[input,rotate=90,minimum width=4.5cm] (input) at (1.25,0) {\large$\textbf{Input spectra}$};
    \node[encoder,rotate=90,minimum width=5.5cm] (conv1) at (3.75,0) {\large$\text{Conv}_{64, 7}$\,+\,$\text{ELU}$\,+\,$\text{BN}$};
    \node[encoder,rotate=90,minimum width=5.5cm] (pool1) at (5,0) {\large$\text{Maxpool}_{2}$};
    \node[encoder,rotate=90,minimum width=5.5cm] (conv2) at (6.25,0) {\large$\text{Conv}_{32, 7}$\,+\,$\text{ELU}$\,+\,$\text{BN}$};
    \node[encoder,rotate=90,minimum width=5.5cm] (pool2) at (7.5,0) {\large$\text{Maxpool}_{2}$};
    \node[encoder,rotate=90,minimum width=5.5cm] (conv3) at (8.75,0) {\large$\text{Conv}_{16, 7}$\,+\,$\text{ELU}$\,+\,$\text{BN}$};
    \node[encoder,rotate=90,minimum width=5.5cm] (pool3) at (10,0) {\large$\text{Avgpool}_{2}$};
    \node[encoder,rotate=90,minimum width=5.5cm] (flat) at (11.25,0) {\large$\text{Flatten}_{224}$};
    
    \node[bottleneck,rotate=90,minimum width=2.5cm] (fc1) at (12.5,-3.125) {\large$\text{Dense}$};
    
    \node[decoder,rotate=90,minimum width=5.5cm] (conv7) at (2.5,-6.25) {\large$\text{Deconv}_{2, 7}$\,+\,$\text{ELU}$};
    \node[decoder,rotate=90,minimum width=5.5cm] (conv6) at (3.75,-6.25) {\large$\text{Deconv}_{64, 7}$\,+\,$\text{ELU}$\,+\,$\text{BN}$};
    \node[decoder,rotate=90,minimum width=5.5cm] (up3) at (5,-6.25) 
    {\large$\text{Upsampling}_{2}$};
    \node[decoder,rotate=90,minimum width=5.5cm] (conv5) at (6.25,-6.25) 
    {\large$\text{Deconv}_{32, 7}$\,+\,$\text{ELU}$\,+\,$\text{BN}$};
    \node[decoder,rotate=90,minimum width=5.5cm] (up2) at (7.5,-6.25) {\large$\text{Upsampling}_{2}$};
    \node[decoder,rotate=90,minimum width=5.5cm] (conv4) at (8.75,-6.25) {\large$\text{Deconv}_{16, 7}$\,+\,$\text{ELU}$\,+\,$\text{BN}$};
    \node[decoder,rotate=90,minimum width=5.5cm] (up1) at (10,-6.25) {\large$\text{Upsampling}_{2}$};
    \node[decoder,rotate=90,minimum width=5.5cm] (fc2) at (11.25,-6.25) {\large$\text{Dense}_{224}$+\,$\text{Reshape}$\,};
    
    \node[output,rotate=90,minimum width=4.5cm] (output) at (1.25,-6.25) {\large$\textbf{Reconstruction}$};
    
    \draw[->] (input) -- (conv1);
    \draw[->] (conv1) -- (pool1);
    \draw[->] (pool1) -- (conv2);
    \draw[->] (conv2) -- (pool2);
    \draw[->] (pool2) -- (conv3);
    \draw[->] (conv3) -- (pool3);
    \draw[->] (pool3) -- (flat);
    
    \draw[-] (flat) -- (12.5,0);
    \draw[->] (12.5,0) -- (fc1);
    \draw[-] (fc1) -- (12.5, -6.25);
    \draw[->] (12.5,-6.25) -- (fc2);
    
    \draw[->] (fc2) -- (up1);
    \draw[->] (up1) -- (conv4);
    \draw[->] (conv4) -- (up2);
    \draw[->] (up2) -- (conv5);
    \draw[->] (conv5) -- (up3);
    \draw[->] (up3) -- (conv6);
    \draw[->] (conv6) -- (conv7);
    \draw[->] (conv7) -- (output);

  \end{tikzpicture}
  }
  \caption{CAE}
  \end{subfigure}
  \captionsetup{font=scriptsize}
  \caption{Model architectures for (a) DAE and (b) CAE. Blue and orange blocks represent the input and output (true and reconstructed spectra) of the models, while the encoder, decoder, and bottleneck are respectively depicted in yellow, pink, and green blocks. At each layer name, one index signifies the shape of the layer, while two indices denote the number of filters and the kernel size.
  }
  \label{fig: model_archs}
\end{figure}
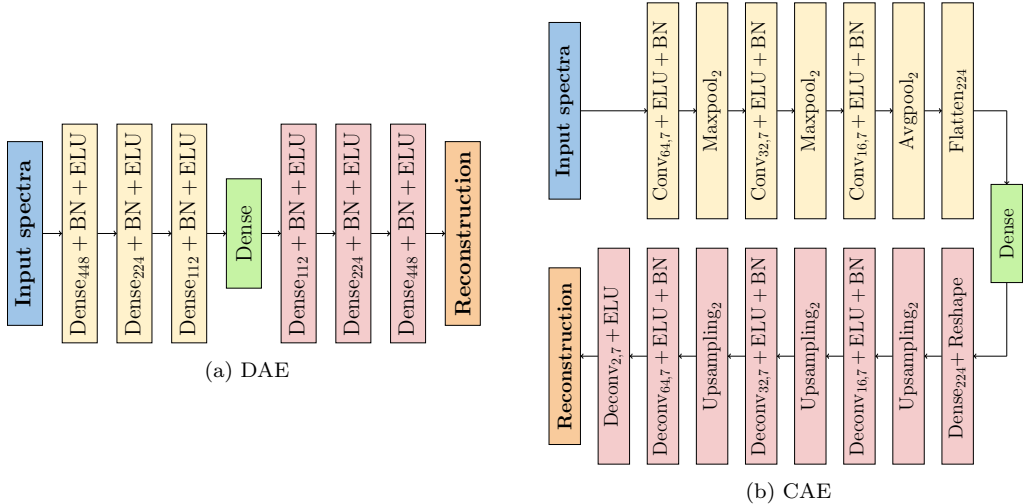

Both our DAE and CAE models maintain a simple architecture. Overviews of the models are provided in Fig. \ref{fig: model_archs}. The DAE comprises a sequence of fully connected dense layers that decrease in size in the encoder part and increase in size in the decoder part. The CAE includes 1D-convolutional layers and max/average pooling layers in the encoder part, with their opposites, transpose, and upsampling layers in the decoder part. The encoder and decoder depths are set to three for each model, as a deeper structure did not yield advantages. Both dense and convolutional layers were augmented with batch normalization \citep[BN;][]{10.5555/3045118.3045167} as the accelerator, and an exponential linear unit \citep[ELU;][]{clevert2016accurate} was applied as the activation function. Other parameters were set by default. Considering the 112 wavelength points and 2 polarimetry parameters, the input and output sizes for our models are both 224 for DAE, with a shape of (112, 2) for CAE. The determination and analysis of the bottleneck (feature vector) size are discussed in Section \ref{subsec_compression_rate_analysis}. The models were implemented using Keras \citep{chollet2015keras} with TensorFlow \citep{tensorflow2015-whitepaper} in the Python \citep{10.5555/1593511}, and the development took place on Google Colaboratory \citep{Bisong2019-fd}.

\subsection{Data preparation}
\label{subsec_data_preparation}
\begin{figure}[htb]
\centering
  \captionsetup{font=scriptsize, justification=centering, margin=0.5cm}
  \includegraphics[width=0.4\linewidth]{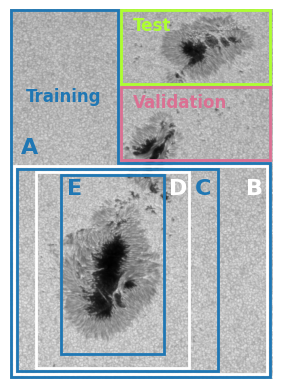}
  \caption{Snapshot of dataset and its partitioning for model training (Versions A to E), validation, and testing.}
  \label{fig:dataset_split}
\end{figure}

The 2D spatial dimension of the selected data has a shape of (512, 722) resulting in a total of 369,664 spectral pixels. The dataset was partitioned into training, validation, and test sets through manual area selection from the spatial image, ensuring the inclusion of both quiet Sun and active regions in all three sets. This resulted in a ratio of approximately 76\% (270,664 px) for training, 12\% (45,000 px) for validation, and 12\% (45,000 px) for test sets. Fig. \ref{fig:dataset_split} displays the dataset split on the continuum image. 

\begin{figure}[htb!]
    \centering
    \captionsetup{font=scriptsize, justification=centering}
    \includegraphics[width=\textwidth]{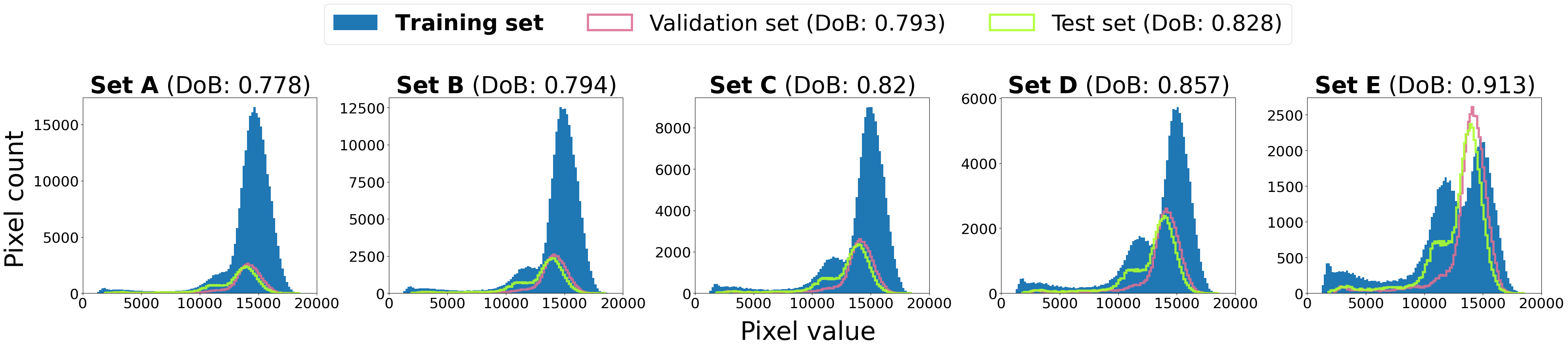}
    \caption{Degree of balance (DoB) in histograms for the five different training sets.}
    \label{fig:degree_of_balance}
\end{figure}

In the training set, the pixel count of the quiet Sun is distinctly higher than that of the sunspot area. Sunspots are uncommon occurrences on the solar surface, leading to their infrequent capture in observational data. Furthermore, the presence of extremely high magnetic fields around and/or within sunspots results in distinct spectral profiles. 
In machine learning, the diversity of data in the training set is important, and balancing the incorporation of different data types during training is essential, as insufficient representation can inhibit the model's ability to effectively learn. This limitation may lead to poor predictions for those specific data types in new datasets. Similarly, if the training set contains a significantly smaller number of sunspot pixels compared to the quiet Sun regions within the spatial image, this could potentially give rise to a data imbalance issue in the training of the deep learning model. Since the data is rarely encountered during training, it is likely to be poorly predicted when it appears in entirely new spectra. To address this issue, we manually prepared five versions of our training set (A (270,664 px), B (216,064 px), C (168,800 px), D (120,000 px), and E (70,000 px)) by considering the ratio of sunspot pixels to quiet Sun pixels. This allows us to explore the impact of data balance on model performances. Consequently, we trained each model five times using these five different training sets, while keeping the validation and test sets the same. Fig. \ref{fig:dataset_split} depicts dataset versions A to E derived from the initial training set.
To quantitatively assess the degree of balance (DoB) for each version of the training set, we calculated the DoB based on the pixel values of the continuum image using Shannon's entropy \citep{shannon1948} method:
\begin{equation}
    {\rm DoB}= \dfrac{-\sum_{i=1}^{k} \dfrac{c_i}{n} \log \dfrac{c_i}{n}}{\log k}, \label{eq:dob}
\end{equation}
where $n$ is the total number of pixels, $k$ is the total number of bins along the pixel value, and $c_i$ is the number of pixels in the $i$-th bin. 
We set $k$ to 100, aiming for a representation of the balance in the training sets that is neither too coarse nor too detailed. The DoB spans from 0 to 1, with a DoB of 0 suggesting unbalanced data and a DoB of 1 indicating balanced data. Fig. \ref{fig:degree_of_balance} shows histograms of the pixel values and their corresponding DoBs.

Utilizing normalized data during the model training enhances both performance and training speed. We applied min-max scaling to normalize the input profile of Stokes I, as 
\begin{equation}
    x'= (b-a)\dfrac{x-\min x}{\max x-\min x}+a \label{eq:minmax},
\end{equation}
where $x$ and $x'$ denote the original and normalized data, respectively, and $[a, b]$ signifies the range for scaling, which for Stokes I is [0, 1].
For Stokes V, we used zero-mean scaling, considering its property of zero centered values, as
\begin{equation}
  x' =
  \begin{cases}
    \dfrac{x}{|\max x|} & \text{if } |\max x| \geq |\min x|\\
    \dfrac{x}{|\min x|} & \text{otherwise}\\
   \end{cases} \label{eq:zeromean}.
\end{equation}

\subsection{Training setups}
\label{subsec_training_setups} 

The models were trained for 1,000 epochs with batch sizes of 512, using the Adam \citep{Kingma2014AdamAM} optimizer. Attempting smaller batch sizes extended the training process excessively, often leading to a halt, with no improvement in performance. We implemented early stopping and learning rate reduction on plateau optimization techniques to enhance the training effectiveness and conserve computational time. The patience parameter for early stopping was set to 100, while for learning rate reduction on plateau, it was set to 50.

In calculating the reconstruction loss function, we computed the mean absolute error (MAE) of intensity values independently for the Stokes I and V parameters at each wavelength point. The total reconstruction loss was then determined by summing the MAE of Stokes I over the MAE of Stokes V, as expressed in
\begin{equation}
    L_{\text{recons}}= {\rm MAE}_{I}+{\rm MAE}_{V}. \label{eq:std_loss}
\end{equation}

\subsection{Evaluation methods} 
\label{subsec_evaluation_methods}

Stokes I features two clearly recognizable absorption lines in the left and right halves of its profile, whereas Stokes V displays four lobes, each pair corresponding to the two absorption line cores. Considering these properties, we defined four target areas for model evaluation based on the root mean square deviation (RMSD) at the respective wavelength ranges—left and right line cores for Stokes I ($LLC_I$, $RLC_I$), and similarly, left and right line cores for Stokes V ($LLC_V$, $RLC_V$). The ranges for the left and right line cores are the same for both parameters: 10–45 and 60–95.

\begin{figure}[htb!]
  \centering
  \captionsetup{font=scriptsize, justification=centering}
  \includegraphics[width=\linewidth]{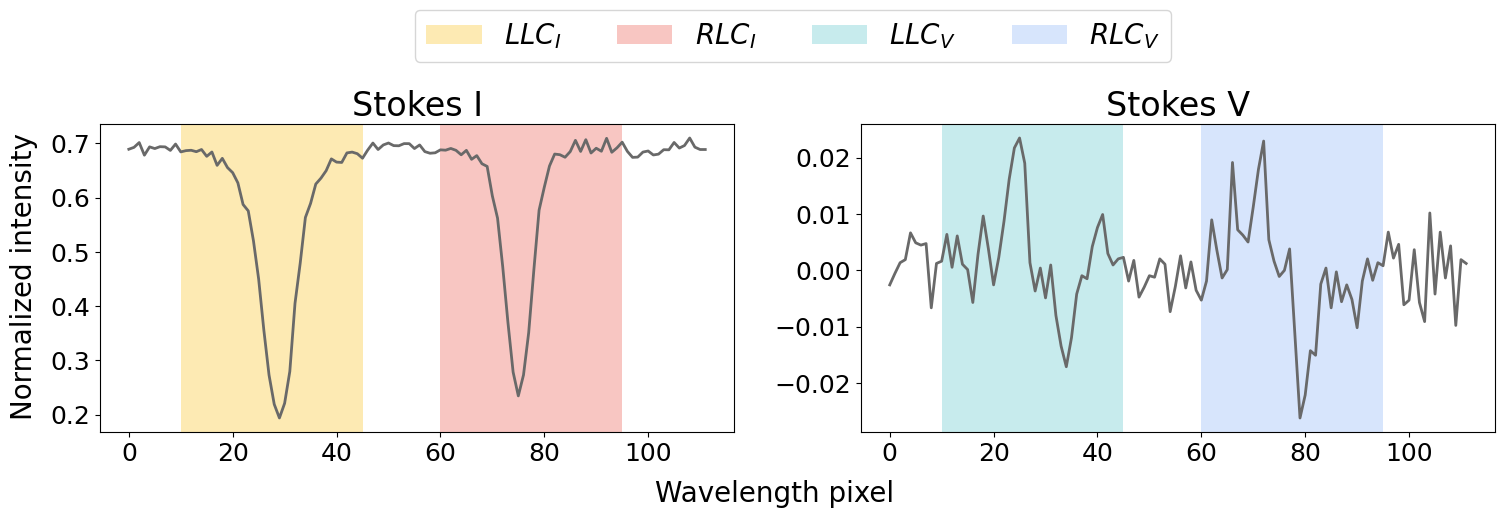}
  \captionof{figure}{Evaluation areas of Stokes profiles. Colored shaded regions indicate the calculations of RMSD for left and right cores of both Stokes parameters ($LLC_{I}$, $RLC_{I}$, $LLC_{V}$, $RLC_{V}$).}
  \label{fig:Stokes_evaluation}
\end{figure}

\section{Results}
\label{sec_results}

\subsection{Model training}
\label{subsec_model_training}

\begin{figure}[htb!]
  \centering
  \captionsetup{font=scriptsize, justification=centering, margin=0cm}
  \includegraphics[width=\linewidth]{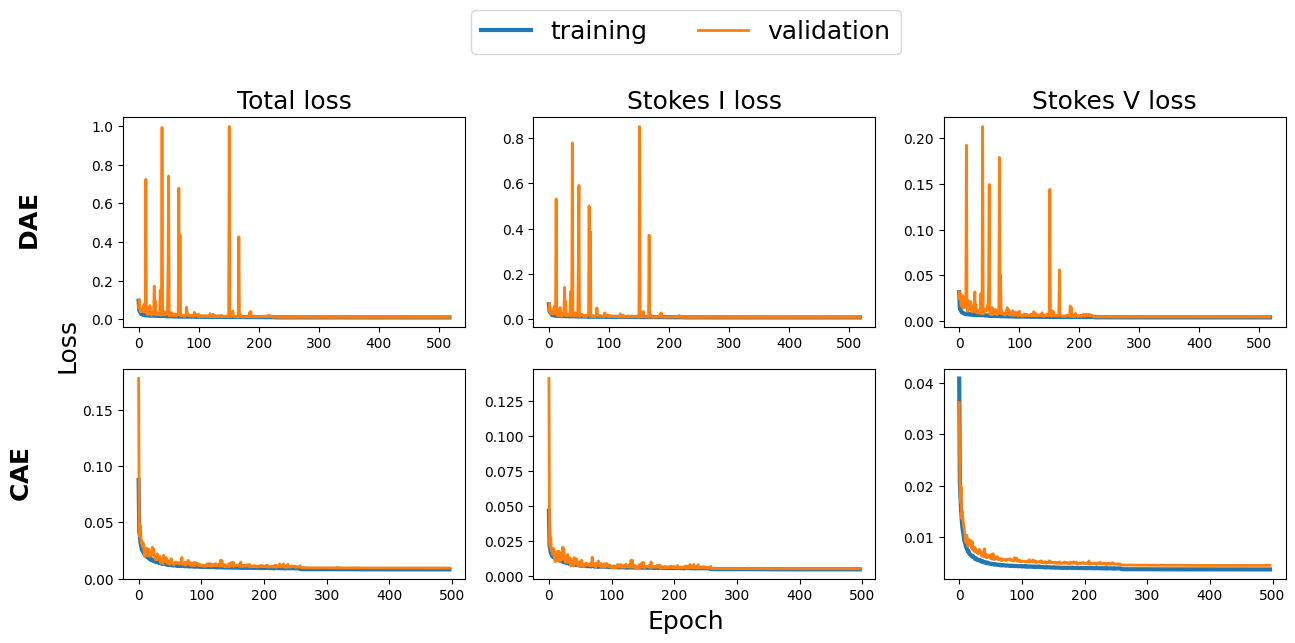}
  \captionof{figure}{Loss-epoch graphs of models trained on set B.}
  \label{fig:loss-epoch}
\end{figure}

We configured the training process with 1000 epochs and implemented early stopping, set to activate if there was no reduction in the reconstruction loss on the validation set for 100 consecutive epochs. The training dynamics are depicted in the loss-epoch dependency graphs for models trained on set B, as shown in Fig. \ref{fig:loss-epoch}. While the DAE model exhibited several sudden jumps in loss during the validation process, the CAE model demonstrates a consistent and smooth decline in validation loss, mirroring the decrease in training loss.

\subsection{Compression rate analysis}
\label{subsec_compression_rate_analysis}

\begin{figure}[htb!]
\centering
  \includegraphics[width=\textwidth]{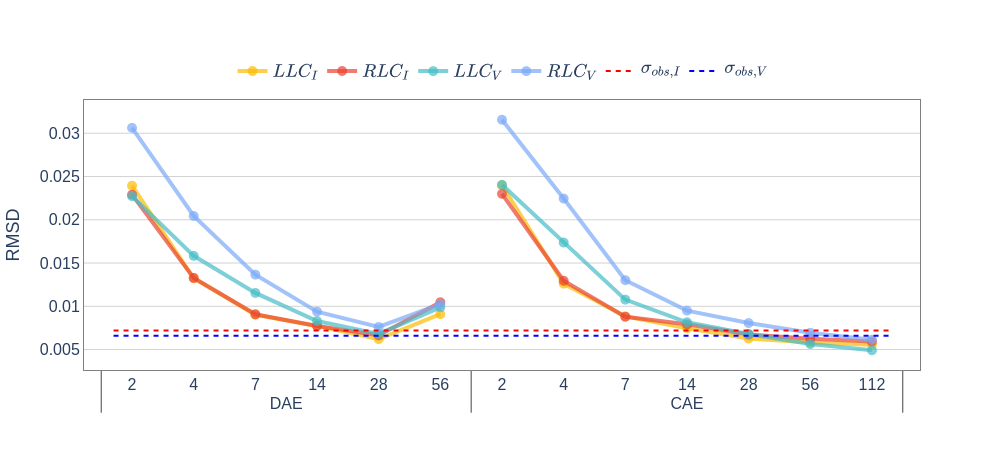}
\captionsetup{font=scriptsize, justification=centering, margin=0.5cm}
\caption{Bottleneck-RMSD dependency graphs of the DAE and CAE models.}
\label{fig:bottleneck_rmsd_dependency}
\end{figure}

To determine the optimal feature vector size for the compression, we conducted experiments by training the models with different bottleneck sizes. The results were analyzed across the four line core areas ($LLC_{I}$, $RLC_{I}$, $LLC_{V}$, $RLC_{V}$), where we compared them with the observational noise levels of Stokes I and V ($\sigma_{obs, I}$, $\sigma_{obs, V}$). For each Stokes parameter, $\sigma_{obs}$ is calculated similarly as
\begin{equation}
\sigma_{obs}=\dfrac{\sum_{i=1}^{N}\sigma^{[0,15]}_i}{N}, \label{eq:std_obs}
\end{equation}
where $N$ represents the number of spectra in the test set, and $\sigma^{[0,15]}_i$ refers to the standard deviation of the continuum within the wavelength range [0, 15] of each spectrum.
The reference continuum level for Stokes V was considered to be 0.
Fig. \ref{fig:bottleneck_rmsd_dependency} shows the dependency between bottleneck size and RMSD values in the target areas of $LLC_{I}$, $RLC_{I}$, $LLC_{V}$, and $RLC_{V}$ for the DAE and CAE models. The horizontal axis indicates the number of nodes in the bottleneck and model types, while the vertical axis displays the RMSD values. The DAE model achieved the lowest RMSDs, approaching the observational noise levels at 28 nodes. However, at 56 nodes, the RMSDs increased, suggesting an overfitting issue. The CAE model also showed a decreasing RMSD trend up to 28 nodes. Notably, its performance continued to improve even at 56 and 112 nodes, with RMSDs falling below the observational noise levels, highlighting the model's robust performance.
Given that 28 nodes in the bottleneck yielded the best possible results for the DAE model, we proceeded by comparing the performances of the DAE and CAE models, both configured with 28 parameters, for spectral reconstruction in the next phase of our analysis.

\subsection{Data imbalance analysis}
\label{subsec_data_balance}

\begin{figure}[htb!]
\centering
  \includegraphics[width=\textwidth]{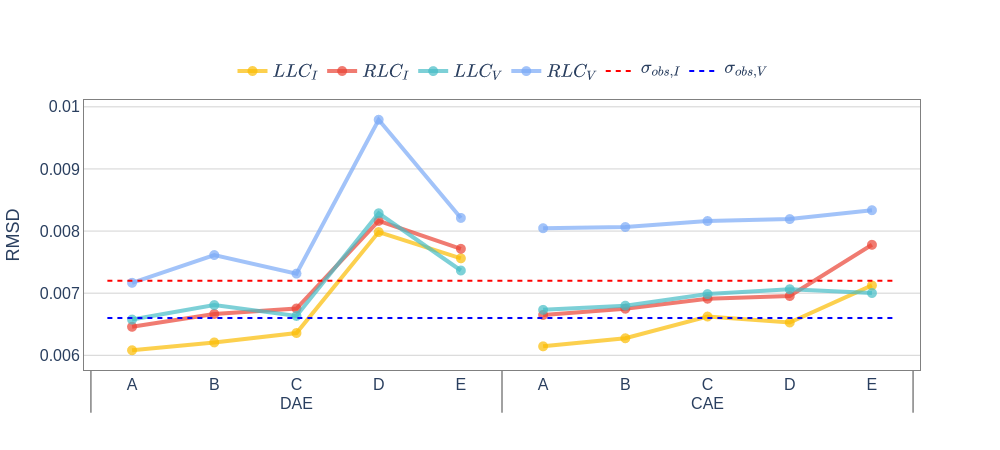}
\captionsetup{font=scriptsize, justification=centering, margin=0.5cm}
\caption{Training set-RMSD dependency graphs of the DAE and CAE models.
}
\label{fig:rmsd_dob_dependency}
\end{figure}

The DAE and CAE models, each having 28 nodes in the bottleneck, were trained using five versions (A to E) of the training sets for evaluations on a common test dataset.
Fig. \ref{fig:rmsd_dob_dependency} illustrates the dependencies between DoB and RMSD values in the target areas. The horizontal axis aligns the training set names and model types, while the vertical axis represents the RMSD values. The DAE exhibited noticeable fluctuations in results, whereas we observed minimal differences in the performance of the CAE. Models trained on sets A, B, and C emerged as top performers, indicating the potential of both models to mitigate training set imbalances.

\subsection{Comparison of observed and reconstructed spectra}
\label{subsec_comparison_of_obs_and_recons}

\begin{figure}[htb!]
\centering
  \includegraphics[width=0.5\textwidth]{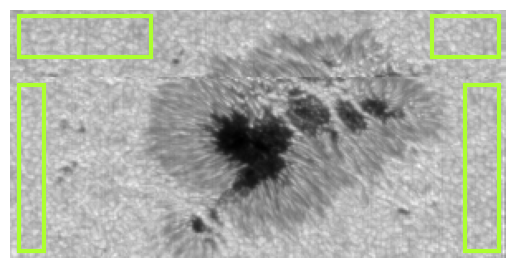}
\captionsetup{font=scriptsize, justification=centering, margin=0.5cm}
\caption{Quiet Sun regions, depicted by green rectangles in the test dataset, are selected to define the continuum intensity of the quiet Sun.}
\label{fig:quietSun_regions}
\end{figure}

\begin{figure}[htb!]
\centering
  \includegraphics[width=\textwidth]{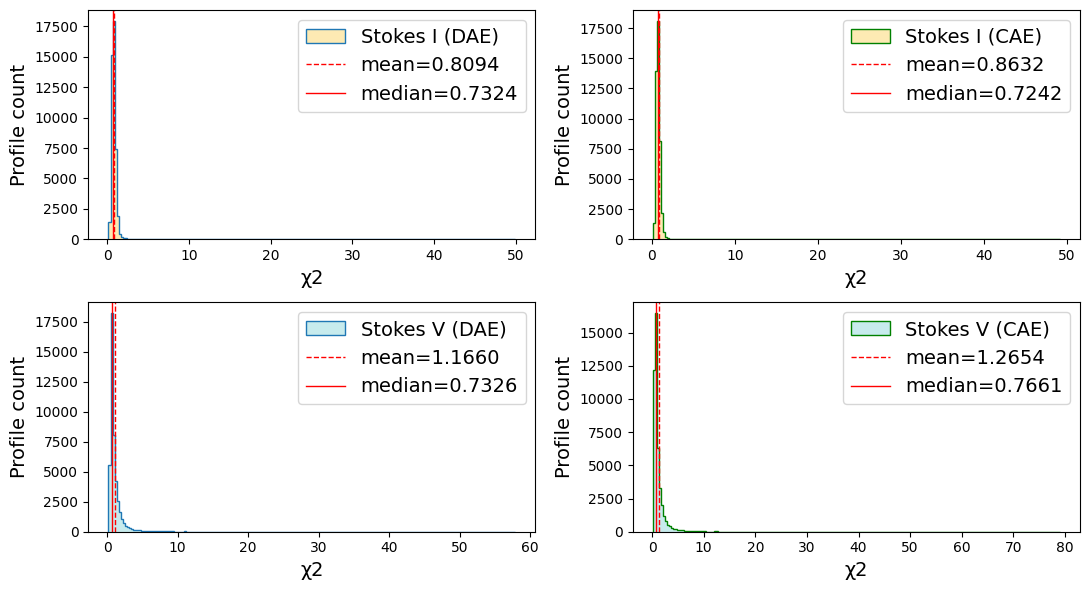}
\captionsetup{font=scriptsize, justification=centering, margin=0.5cm}
\caption{Chi-square histograms of observed and reconstructed spectra.
}
\label{fig:chisq_histogram}
\end{figure}

\begin{figure}[htb!]
\centering
  \includegraphics[width=\textwidth]{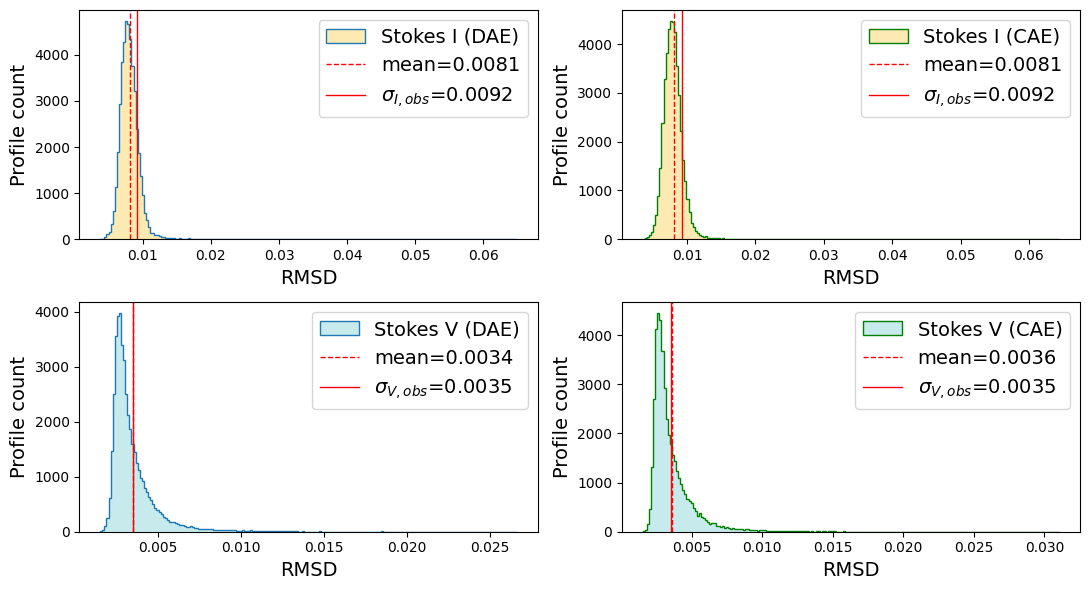}
\captionsetup{font=scriptsize, justification=centering, margin=0.5cm}
\caption{RMSD histograms of observed and reconstructed spectra.
}
\label{fig:rmsd_histogram}
\end{figure}

We compared the original and reconstructed profiles from the models with the best performances based on chi-squared and RMSD metrics. The reconstructed profiles are unscaled to original intensity range from the scaled model output. The observational test profiles with original intensity and the unscaled reconstructions are then normalized to the quiet Sun continuum ($I_c$). Quiet Sun regions selected from the test set, as shown in Fig. \ref{fig:quietSun_regions}, are used to define $I_c$.
Chi-square values ($\chi^2$) are calculated wavelength-wise with respect to $\sigma_{obs}$ for each profile as
\begin{equation}
    \chi^2=\dfrac{1}{d}\sum_{i=1}^{d}\dfrac{(S'(\lambda_{i})-S(\lambda_{i}))^2}{\sigma^2_{obs}},
    \label{eq:chisq_IV}
\end{equation}
where $d$ represents the number of data points in the continuum within the wavelength range [0, 15], and $S(\lambda_{i})$ and $S'(\lambda_{i})$ refer to the spectral intensity at $i$-th wavelength point of observational and reconstructed spectra, respectively. Fig. \ref{fig:chisq_histogram} shows the chi-square histograms of the distribution across all test data for Stokes I and V, comparing the performances of the DAE and CAE models. Stokes I histograms resulted in mean values less than 1, suggesting that the average $\chi^2$ values are within the observational noise. For Stokes V, both models yielded mean values slightly above 1, particularly in the case of CAE.
Similarly, RMSD histograms of Stokes I and V reconstructions obtained from the DAE and CAE models are depicted in Fig. \ref{fig:rmsd_histogram}, using the calculation for each profile as
\begin{equation}
    {\rm RMSD}=\sqrt{\dfrac{\sum_{i=1}^{d}(S'(\lambda_{i})-S(\lambda_{i}))^2}{d}},
    \label{eq:rmsd_IV}
\end{equation}
suggesting that all histograms resulted in mean RMSD values comparable to the observational noise.

\begin{figure}[htb!]
  \centering
  \captionsetup{font=scriptsize, justification=centering, margin=1cm}
  \includegraphics[width=0.81\linewidth]{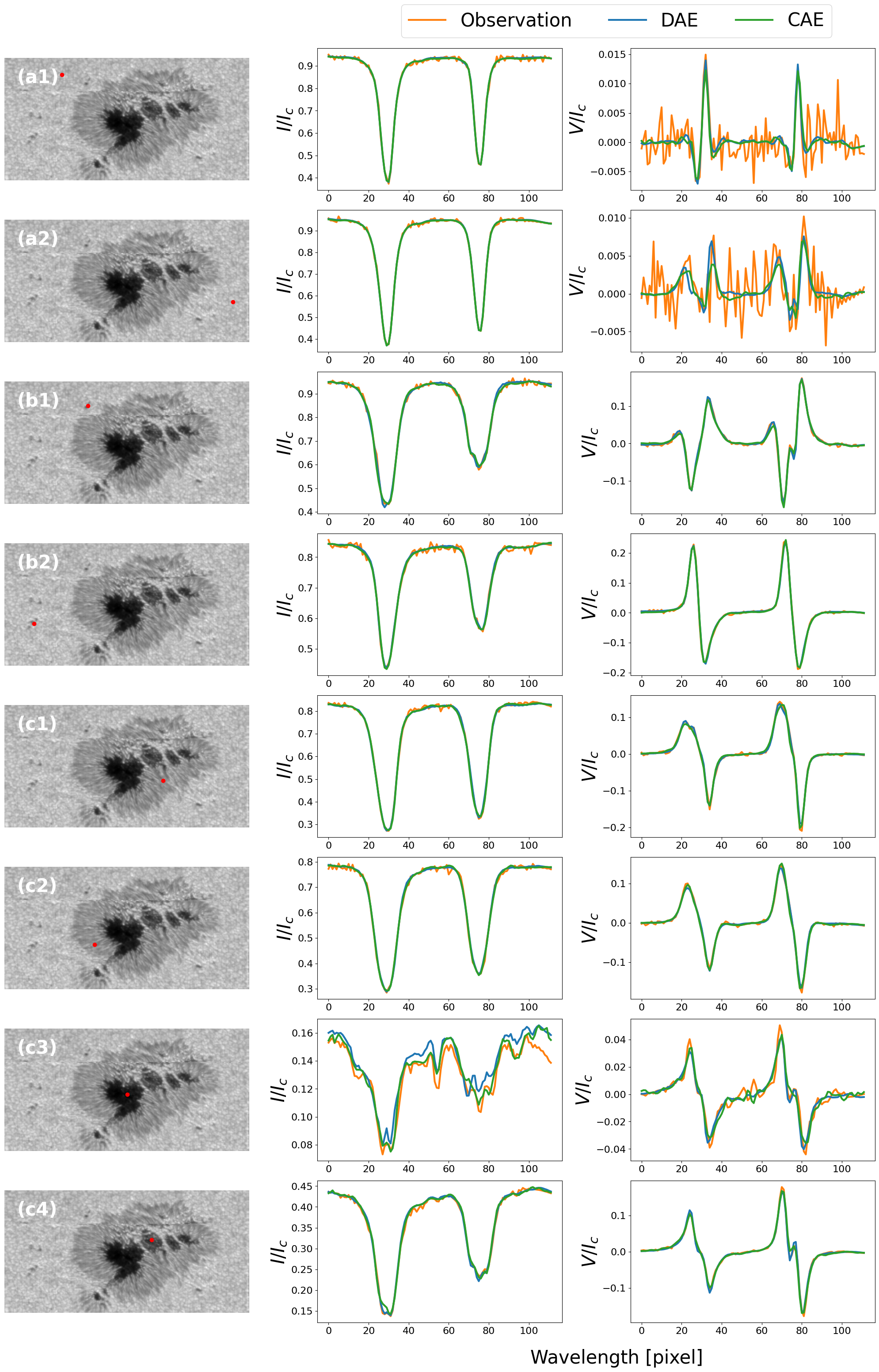}
  \captionof{figure}{Samples of observational and reconstructed spectra at various spatial positions, including (a1–2) quiet Sun, (b1–2) pore, and (c1–4) sunspot penumbra and umbra, in the test set.}
  \label{fig:reconstruction_samples}
\end{figure}

To display reconstruction samples, we chose eight different spatial positions, including the quiet Sun, pores, and both the penumbra and umbra of a sunspot in the continuum image of the test set. Figure \ref{fig:reconstruction_samples} illustrates comparisons between the true observational profiles of selected pixel positions and their respective reconstructions from the DAE and CAE models.
Overall, both models produced smooth and comparable reconstructions that accurately fit the entire profiles, including fluctuations such as Stokes I continuums. Importantly, the reconstructions effectively captured Stokes V shapes in the quiet Sun despite the initial high noise levels, achieving a good balance of noise removal without overfitting. In high magnetic field regions like the sunspot center in (c3), where the Stokes profiles are rarer and more complex, both models faced challenges in accurate reconstruction, with the DAE model showing particular difficulty.

\clearpage

\section{Discussion}
\label{sec_discussion}

To achieve the highest compression rate possible while ensuring robust performance relative to observational noises, we opted to use 28 nodes in the bottleneck for both the DAE and CAE models. A larger bottleneck in the CAE model has the potential to further improve its ability to reconstruct spectral shapes.

Stokes V profiles typically exhibit higher and less clearly definable noise levels compared to Stokes I profiles, which prompted concern about their potential impact on the training process and the risk of undertraining Stokes I signals. Importantly, we found that optimizing the balance between the contributions of Stokes I and Stokes V in the reconstruction loss during training—such as by customizing the loss function with a specific weight for the Stokes V component—was unnecessary to achieve satisfactory model performance.

When using different datasets for training, the DAE model exhibited a higher sensitivity to the DoB in the training set. It faced challenges in accurately reconstructing spectral profiles that were less presented during training. In contrast, the CAE models demonstrated greater flexibility in this regard, consistently delivering stable performances across varying DoB in the training dataset.

The novel aspect of this work lies in the integrated compression of two polarimetric parameter profiles, applied not only to the quiet Sun but also to various positions on the solar surface, including active regions. This approach enables the analysis of spectral profiles in regions of interest with strong magnetic fields, which could potentially drive a range of solar behaviors. The study is limited by its reliance on only the Stokes I and V parameters from the four Stokes spectra. To address this, future research should incorporate the Stokes Q and U parameters, which will provide a more comprehensive understanding of the solar atmosphere's structure, physical conditions, and complex magnetic fields. Additionally, while our current analysis is confined to data from the disk center, future studies should extend this focus to encompass other regions across the solar disc.

Our compression method shows potential for a wide range of applications, such as detection of anomalous spectra. In this scenario, the model is training on normal data to ensure a reconstruction error below a specified threshold. Subsequently, the pre-trained model is applied to reconstructing data containing previously unseen anomalous signals, resulting in a reconstruction error that exceeds the threshold. Furthermore, unusual events, such as solar flares, could possibly be detected based on their distinctive spectral signatures observed in data preceding actual flare occurrences. Additionally, the suggested approach potentially facilitates the comparison and analysis of observational and numerical simulation data by leveraging the compact representations provided by the compression model.

\section{Conclusion}
\label{sec_conclusion}

In this work, we developed two distinct deep learning model architectures, a deep autoencoder (DAE) and a 1D-convolutional autoencoder (CAE), specifically tailored for compressing Hinode SOT/SP spectral data, with a primary focus on the Stokes I and V polarization parameters. Our experiments aimed to determine the optimal compression rate, evaluate different model architectures, and assess their performance across various balanced training datasets.

The results demonstrate that our compression models effectively reduced the spectral data dimensionality from 224 to 28 parameters, yielding reconstruction residuals comparable to the observational noise while also eliminating high noise levels. Notably, the CAE model outperformed the DAE model, offering greater stability in handling data imbalance and robustly maintaining the reproducibility of complex profile shapes.

The novelty of our study lies in the compression of two-dimensional observational solar polarimetric spectra in both the quiet Sun and active regions. This method provides a more effective analysis technique, significantly broadening its applicability for solar physics studies.

In future work, we aim to develop a universal compression model that improves detailed spectral analysis by incorporating full Stokes parameters and can be applied to a broad range of snapshots, extending beyond the disk center.

\section*{Acknowledgement}
\label{sec_acknowledgement}
We would like to acknowledge the reviewers for their detailed and thoughtful comments. Their expertise and suggestions were instrumental in greatly improving our results.

This work was supported by JST  Grant Number JPMJFS2114 as a university fellowship towards the creation of science technology innovation.

Hinode is a Japanese mission developed and launched by ISAS/JAXA, collaborating with NAOJ as a domestic partner and with NASA and STFC (UK) as international partners. Scientific operation of the Hinode mission is conducted by the Hinode science team organized at ISAS/JAXA. This team mainly consists of scientists from institutes in the partner countries. Support for the post-launch operation is provided by JAXA and NAOJ(Japan), STFC (U.K.), NASA, ESA, and NSC (Norway).

\bibliographystyle{elsarticle-harv} 
\bibliography{bibliography}

\end{document}